\documentclass{tmsce}

\usepackage{amsmath,amsfonts,amsthm}

\usepackage{algorithm}
\usepackage{array}
\usepackage[caption=false,font=normalsize,labelfont=sf,textfont=sf]{subfig}
\usepackage{textcomp}

\usepackage{url}
\usepackage{verbatim}
\usepackage{graphicx}
\usepackage{cite}

\usepackage{float} 
\usepackage{algpseudocode} 
\usepackage{color} 

\fancypagestyle{firstpage}{
  \fancyhf{}
  \fancyhead[R]{\small $\dagger$ Equal contribution}
  \fancyfoot[C]{}
  \fancyfoot[R]{\small \thepage}

}

\definecolor{Gray}{gray}{.25} 

\vol{}
\issue{}
\yearofpub{}
\pagerange{}

\title{Inference-Time Loss-Guided Colour Preservation in Diffusion Sampling}

\authors{Aarush Ram Anandh$^{1,\dagger}$ and Angad Singh Ahuja$^{1,\dagger}$}
\affiliation{$^{1}$Constrained Image-Synthesis Lab}


\begin{document}
\maketitle
\thispagestyle{firstpage}

\begin{abstract}
Precise color control remains a persistent failure mode in text-to-image diffusion systems, particularly in design-oriented workflows where outputs must satisfy explicit, user-specified color targets. We present an inference-time, region-constrained color preservation method that steers a pretrained diffusion model without any additional training. Our approach combines (i) ROI-based inpainting for spatial selectivity, (ii) background-latent re-imposition to prevent color drift outside the ROI, and (iii) latent nudging via gradient guidance using a composite loss defined in CIE Lab and linear RGB. The loss is constructed to control not only the mean ROI color but also the tail of the pixelwise error distribution through CVaR-style and soft-maximum penalties, with a late-start gate and a time-dependent schedule to stabilize guidance across denoising steps. We show that mean-only baselines can satisfy average color constraints while producing perceptually salient local failures, motivating our distribution-aware objective. The resulting method provides a practical, training-free mechanism for targeted color adherence that can be integrated into standard Stable Diffusion inpainting pipelines.
\end{abstract}

\printkeywords

\tmsceendfrontmatter

\section*{Introduction and Motivation}

Diffusion models have become a standard backbone for high-quality image synthesis and editing, but their controllability is still largely mediated by semantic conditioning (text) and structural conditioning (edges, poses, segmentation, inpainting masks). While these mechanisms are effective for composition and geometry, they remain unreliable for enforcing low-level numeric attributes such as exact color targets within a specified region of interest (ROI). In practical design workflows, however, color is often specified as an explicit constraint (e.g., a brand palette entry, a UI theme color, a packaging swatch), and the requirement is not merely that an output is ``roughly blue,'' but that it matches a given sRGB or perceptual color encoding within an acceptable tolerance and without unintended drift across the rest of the image.

This paper studies inference-time color preservation as a constrained sampling problem. We treat the diffusion sampling trajectory as the control surface and apply lightweight guidance signals directly to the latent state during denoising.  Our loss design mixes two complementary representations: a decaying linear-RGB term that biases early denoising toward the target chromatic direction, and a progressively strengthened Lab-space, tail-sensitive term that penalizes persistent outliers as structure becomes coherent. This yields a modular inference-time control mechanism that does not require retraining, does not introduce new parameters, and can be plugged into existing Stable Diffusion inpainting pipelines.

\medskip

A core motivation is to prioritize controllability under compute and integration constraints. Many approaches to improving visual fidelity under constraints (e.g., texture realism, high-frequency detail preservation, or learned region-specific appearance priors) require either additional training, auxiliary networks, or repeated inner-loop optimization that materially increases inference cost and system complexity. In contrast, our objective is narrower: we seek predictable, ROI-local color adherence with minimal disruption to the base model's generative prior. The proposed method therefore avoids explicit texture or patch-level appearance matching losses. The method is intended as a lightweight, optimized baseline for color-critical editing tasks, and as a foundation on top of which richer appearance constraints (including texture-aware or print-aware objectives) can later be layered when the additional complexity is warranted.

\medskip

\section*{Literature Review}

Diffusion models fundamentally establish image generation as a sequential noising-and-denoising process. Ho et al. \cite{ho2020ddpm} introduced Denoising Diffusion Probabilistic Models, which learn to reverse a gradual Gaussian noising of data to produce a high-fidelity image. The formulation that they use focuses on optimising a variational upper-bound, connected to score matching with the following form - 

\begin{equation}
q(\mathbf{x}_t \mid \mathbf{x}_0) \;=\; \mathcal{N}\!\Big(\mathbf{x}_t;\sqrt{\bar{\alpha}_t}\,\mathbf{x}_0,\,(1-\bar{\alpha}_t)\mathbf{I}\Big)
\quad\Longleftrightarrow\quad
\mathbf{x}_t \;=\; \sqrt{\bar{\alpha}_t}\,\mathbf{x}_0 \;+\; \sqrt{1-\bar{\alpha}_t}\,\boldsymbol{\epsilon},\;\;\boldsymbol{\epsilon}\sim\mathcal{N}(\mathbf{0},\mathbf{I}).
\end{equation}
\begin{equation}
\mathcal{L}_{\text{simple}}(\theta)
\;=\;
\mathbb{E}_{t,\mathbf{x}_0,\boldsymbol{\epsilon}}
\Big[\big\|\boldsymbol{\epsilon}-\boldsymbol{\epsilon}_{\theta}(\mathbf{x}_t,t)\big\|_2^2\Big].
\end{equation}
\medskip

Nichol and Dhariwal \cite{nichol2021improvedddpm} refined this paradigm significantly by learning noise variance and adjusting the diffusion schedule allowing for accelerated sampling by orders of magnitude with negligible quality loss. Simultaneously, Song et al. \cite{song2020ddim} proposed DDIM which reuses the DDPM training methodology but used a non-markovian sampler allowing for deterministic or fewer-step generation with only minor degradation. Song et al. \cite{song2020sde} finally unified this theoretical foundation by construction diffusion as a Stochastic Differential Equation such that data is continuously noised to a prior, and a learned time-dependent score function defines the reverse SDE to sample new data in the following formulation.

\begin{equation}
d\mathbf{x} \;=\; \mathbf{f}(\mathbf{x},t)\,dt \;+\; g(t)\,d\mathbf{w}_t,
\end{equation}
\begin{equation}
d\mathbf{x} \;=\; \Big[\mathbf{f}(\mathbf{x},t)\;-\;g(t)^2\,\nabla_{\mathbf{x}}\log p_t(\mathbf{x})\Big]\,dt \;+\; g(t)\,d\bar{\mathbf{w}}_t.
\end{equation}
\medskip

This allowed for new sampling procedures and test-time constraints. For our purposes, It shows how one can inject an additional energy or guidance term into the generative SDE to steer outputs. These developments establish that the diffusion sampling process itself provides hooks to modulate generation which we intend to use.  

\medskip

Early work on controllable diffusion introduced guidance strategies that add explicit gradients during sampling. Dhariwal \& Nichol \cite{dhariwal2021diffusionbeatgans} first showed that diffusion models can surpass GANs in image quality and diversity, and they proposed classifier guidance as a way to trade off diversity for fidelity. Classifier guidance demonstrated the principle of test-time steering: one can inject an arbitrary differentiable goal into the generation dynamics without retraining the generative model.  Ho \& Salimans \cite{ho2022cfg} later generalized this to classifier-free guidance (CFG). CFG became the de facto standard in modern text-to-image diffusion (as used in Stable Diffusion, GLIDE \cite{nichol2021glide}, etc.) because it is simple and effective: it yields superior fidelity to prompt details compared to unguided samples. However, CFG’s control is implicit and global meaning that it cannot guarantee adherence to specific attributes like exact color values, which often remain “emergent” behaviors of the model. As a result, while one can roughly ask for “a blue car” and get a blue-ish car, one cannot precisely constrain the shade or prevent the model from introducing off-palette tones. This limitation motivates more structured control mechanisms as are necessarily needed for industrial and design use-cases.

\medskip

Researchers have incorporated stronger conditioning streams and architectural modifications to tame diffusion outputs. An example would be Latent Diffusion Models (LDM) \cite{rombach2021ldm}, as introduced by Rombach et al., which move the diffusion process into a lower-dimensional latent space learned by an autoencoder. 

\begin{equation}
\mathbf{z}=E(\mathbf{x}),\quad 
\mathbf{z}_t \;=\; \sqrt{\bar{\alpha}_t}\,\mathbf{z} \;+\; \sqrt{1-\bar{\alpha}_t}\,\boldsymbol{\epsilon},\;\;\boldsymbol{\epsilon}\sim\mathcal{N}(\mathbf{0},\mathbf{I}),
\quad
\mathcal{L}_{\text{LDM}}(\theta)
\;=\;
\mathbb{E}_{t,\mathbf{z},\boldsymbol{\epsilon}}
\Big[\big\|\boldsymbol{\epsilon}-\boldsymbol{\epsilon}_{\theta}(\mathbf{z}_t,t,\mathbf{c})\big\|_2^2\Big].
\end{equation}
\medskip

Rombach et al. \cite{rombach2021ldm} equip the diffusion model with a cross-attention mechanism to condition on text embeddings, typically from a pretrained language-image model like CLIP \cite{radford2021clip} or BERT \cite{devlin2019bert}. Yet as Radford et al.’s CLIP model is oriented toward overall semantic alignment \cite{radford2021clip}, certain low-level attributes (such as precise color tones or exact geometry) are not reliably or directly specified by text alone \cite{butt2025gencolorbench}. In practice, current text-to-image models often struggle with prompts specifying uncommon or very specific colors (e.g. “a chartreuse handbag with Pantone 448 C straps”) \cite{butt2025gencolorbench}. The color words in the prompt guide the general hue, but the model might drift toward more typical colors or introduce unintended color blends, because the language conditioning is learned to satisfy human-describable semantics rather than enforce pixel-level constraints. Thus, purely text-based color control is at best indirect. Recent evaluations confirm this as is seen in a comprehensive benchmark (GenColorBench) (Butt et al. 2025) \cite{butt2025gencolorbench} with 44k color-focused prompts covering over 400 named or numeric colors, SOTA diffusion models showed significant errors in color accuracy and consistency.

\medskip

Researchers have therefore developed specialized conditioning mechanisms for other image properties, specifically structural or spatial controls, that could serve as inspiration for color control. For example, ControlNet (Zhang et al.) \cite{zhang2023controlnet} augments a frozen diffusion model with a trainable branch that takes an extra condition image (such as an edge map, human pose skeleton, segmentation map, etc.) . This branch is attached via zero-convolution layers to the main model and is trained to modulate the denoising process in accordance with the conditioning image. Notably, the base model’s weights remain untouched (avoiding catastrophic forgetting), and multiple conditions can be combined. Works on structural and spatial control underscore a pattern, which is that adding targeted conditioning pathways can significantly improve controllability. Color control can be seen in the same light instead of relying on the diffuse influence of a text encoder, one can imagine a dedicated conditioning channel for color attributes (e.g. a palette or target histogram) that steers generation more reliably. Indeed, a recent survey by Pu et al. categorizes controllable generation methods along these lines \cite{cao2024controllablesurvey}, noting that current systems handle spatial and structural conditions much better than they handle precise color or numeric conditions.

\medskip

Explicit color control in diffusion models has only very recently seen focused attention, and remains limited. One line of work conditions generation on a given color palette or distribution. Vavilala et al. (2023) \cite{vavilala2023palette_local_control} pioneered a basic palette-conditioning approach, showing that a standard diffusion model can roughly obey a provided palette by concatenating palette information to the model input, albeit with only coarse control. A more formal advance is by Aharoni et al., who propose Palette-Aligned Diffusion with a Palette-Adapter module \cite{aharoni2025palette_aligned}. In their approach, the user supplies a set of reference colors, and the model is guided to use exactly those colors (and no others) in the output. They interpret the palette as a target color histogram and introduce two scalar parameters to control how strictly the model adheres to that. By training on a broad dataset of images with known color distributions, their adapter learns to “align” the diffusion generation to the palette, yielding significantly improved color fidelity over naive prompt-based control. This is an important proof that diffusion models can internalize hard color constraints when given proper conditioning mechanisms. Another relevant work is by Agarwal et al., who explore a training-free technique for color and style disentanglement \cite{agarwal2024trainingfree_colorstyle}. Rather than fine-tuning the model, they operate on its latent features at inference by adjusting the feature covariance to match a reference image’s color statistics and manipulate self-attention on luminance vs. chrominance channels to separate style from color.  Notably, this is done without additional training as it is an optimization applied per image. This underscores the promise of test-time optimization for color control: if one can compute some differentiable measure of color similarity, one can plug it into the diffusion process directly. Overall, these works hint that direct color control is achievable, but either require additional training (palette adapter) or only weakly enforce color (post-hoc latent tweaks). Crucially, none yet combine color constraints with spatial selectivity (ensuring the right regions get the right colors). Typically they affect the whole image’s palette. Our approach intends to fill this gap by allowing region-based color constraints at test time. 

\medskip

In designing our solution, we draw inspiration from the successes in constrained sampling for diffusion models, as established in the diffusion literature on inverse problems. A core idea is that you can introduce an additional term during each sampling step that “guides” the sample toward satisfying some constraint, effectively performing a Lagrangian optimization on the fly. Chung et al.’s Diffusion Posterior Sampling (DPS) \cite{chung2022dps} is an approach to solve a noisy inverse problem , they augment each diffusion iteration with the gradient as specified below - 

\begin{equation}
\nabla_{\mathbf{x}_t}\log p(\mathbf{x}_t\mid \mathbf{y})
\;\approx\;
\mathbf{s}_{\theta}(\mathbf{x}_t,t)
\;+\;
\nabla_{\mathbf{x}_t}\log p\!\Big(\mathbf{y}\mid \hat{\mathbf{x}}_0(\mathbf{x}_t)\Big),
\end{equation}
\begin{equation}
\hat{\mathbf{x}}_0(\mathbf{x}_t)
\;=\;
\frac{1}{\sqrt{\bar{\alpha}_t}}
\Big(\mathbf{x}_t-\sqrt{1-\bar{\alpha}_t}\,\boldsymbol{\epsilon}_{\theta}(\mathbf{x}_t,t)\Big).
\end{equation}
\medskip

This gradually steers the denoising trajectory into the subset of images consistent with the observations. Importantly, DPS requires no retraining of the diffusion model; it operates as a test-time wrapper. The resulting sampler blends the original generative score with a “manifold constraint” term, yielding a more accurate and consistent solution than unguided diffusion \cite{chung2022dps}. Follow-up works like DDRM (Kawar et al. 2022) \cite{kawar2022ddrm} and Wang et al. (DDNM) \cite{wang2022ddnm} refine this by characterizing the null-space of the measurement and ensuring the diffusion updates only affect the parts of the image that don’t contradict. The broader point is that adding a custom energy or loss term to the diffusion sampling procedure is a powerful and principled way to impose constraints \cite{chung2022dps}. Recent surveys echo that test-time conditioning is now a standard approach for avoiding expensive model retraining when new conditions are desired \cite{cao2024controllablesurvey}. In our case, the “measurement” is the set of color constraints (e.g. a target histogram or target Lab values in an ROI), and we incorporate their penalty gradients into the diffusion updates similarly. By doing so, we inherit a key advantage: we do not need to fine-tune the model’s weights to enforce color – we only adjust the sampling path, which avoids any risk of degrading the model’s generative prior or requiring large training datasets of color-specific examples. This strategy is similar with recent work by Lobashev et al., as is proposed as Sliced-Wasserstein (SW) guidance for color distribution alignment \cite{lobashev2025sw_guidance}. They define a differentiable loss between the generated image’s color distribution and a reference distribution (using a sliced Wasserstein distance on Lab histograms), and backpropagate that at each step which is specified below -

\begin{equation}
\mathcal{L}_{\text{SW}}(\mathbf{x})
\;=\;
\mathrm{SW}_p\!\Big(\mu(\phi(\mathbf{x})),\,\mu(\phi(\mathbf{x}^{\text{ref}}))\Big),
\qquad
\mathrm{SW}_p(\mu,\nu)
\;=\;
\int_{\mathbb{S}^{d-1}}
W_p\!\Big((\mathcal{P}_{\boldsymbol{\theta}})_{\#}\mu,\;(\mathcal{P}_{\boldsymbol{\theta}})_{\#}\nu\Big)\,d\boldsymbol{\theta}.
\end{equation}
\medskip

This method, applied to text-to-image diffusion, yielded images that very closely match a reference palette’s overall color usage while still adhering to the text prompt’s semantics. Crucially, no model training was needed. The guidance operates purely at sampling time. Our approach can be seen as extending this idea, but with more granular control (specific regions and multiple loss terms to balance fidelity and quality).

\medskip

\medskip

\subsection*{Color Conversions for Inference-Time Losses}
To compute the inference-time losses described in this paper, we require consistent and differentiable conversions between multiple color encodings. Our composite objective is a weighted sum of sub-losses defined in (i) CIE $L^{*}a^{*}b^{*}$ (Lab) space and (ii) linear-light RGB space. The user provides a single target color as a normalized sRGB triplet in $[0,1]^3$, which we denote as $\mathrm{sRGB}_{\text{user}}$. Because sRGB is gamma-compressed, it is not equivalent to linear RGB; therefore, when a sub-loss is defined in linear RGB, we first apply the standard sRGB inverse transfer function to obtain $\mathrm{LRGB}_{\text{user}}$ (linear-light RGB). We additionally convert the user color into Lab, denoted $\mathrm{Lab}_{\text{user}}$, using a differentiable sRGB$\rightarrow$Lab conversion consistent with the CIE 1976 definition of Lab.

At inference time, our losses act on the evolving sample produced by the diffusion sampler. At each timestep $t$, we decode the current latent into an image in the model’s RGB output space and clamp it to $[0,1]$ to obtain a normalized sRGB image $\mathrm{sRGB}_{t}$. From this, we compute (a) the corresponding linear-light RGB image $\mathrm{LRGB}_{t}$ via sRGB linearization, and (b) the corresponding Lab image $\mathrm{Lab}_{t}$ via the standard Lab conversion. We thus maintain four primary quantities used by the loss terms:
(i) $\mathrm{LRGB}_{\text{user}}$, (ii) $\mathrm{Lab}_{\text{user}}$, (iii) $\mathrm{LRGB}_{t}$, and (iv) $\mathrm{Lab}_{t}$.
Lab-based losses are defined as discrepancies between $\mathrm{Lab}_{t}$ and $\mathrm{Lab}_{\text{user}}$ (typically within a specified ROI), while linear-RGB-based losses are defined as discrepancies between $\mathrm{LRGB}_{t}$ and $\mathrm{LRGB}_{\text{user}}$. All conversions are implemented with differentiable operators to permit backpropagation of the loss signal to the latent during sampling \cite{riba2020kornia,cie11664_4_2019,icc_srgb}.

\medskip

\subsection*{Conversion Pipeline}

\begin{figure}[H]
\centering
\includegraphics[width=\linewidth]{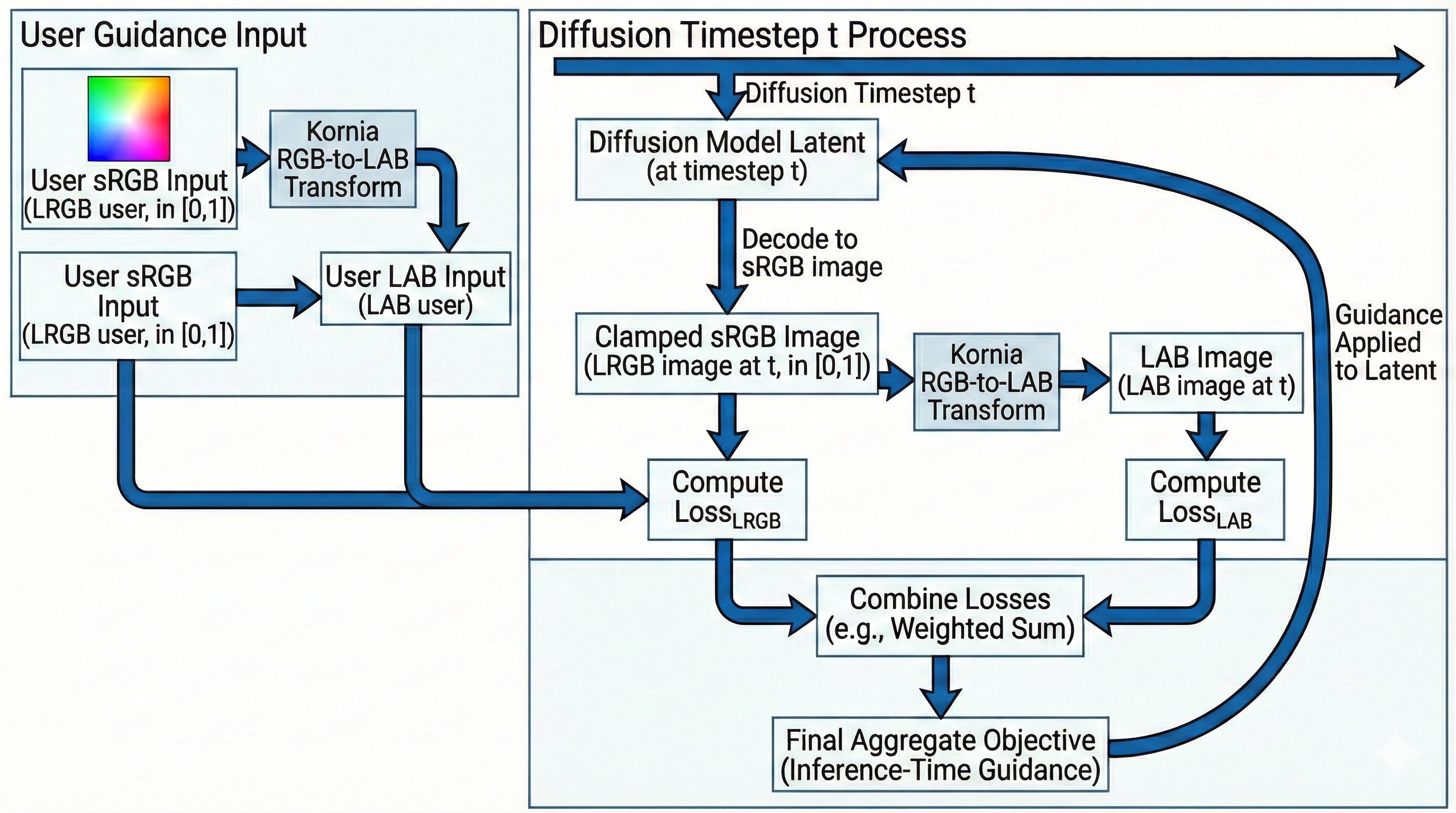}
\caption{\textbf{Conversion pipeline.} Overview of the differentiable colour conversion and loss computation flow.}
\label{fig:pipeline}
\end{figure}

\section*{Loss Functions and Inpainting Procedure}

We begin with the assumption that there exists a single master loss function applied at inference time. This loss takes as inputs the user-specified targets, the model output at timestep $t$, and a small set of weights and hyperparameters. During the diffusion process, a latent representation is generated at each step. At each timestep $t$, we decode the latent and perform the color-space conversions described previously, after which the master loss $L_t$ is computed for the latent at timestep $t$.

\medskip

To enforce and apply these losses, we use an inference-time guidance strategy referred to here as latent nudging. This can be viewed as a corrective push applied at each timestep by taking a gradient step on the latent with respect to the loss. Since the loss is used as a guidance signal to locally bias the denoising trajectory, it is desirable that the effective loss signal does not vary too abruptly between successive timesteps. The nudging update is given by:
\begin{equation}
z_t \leftarrow z_t - \eta \nabla_{z_t} L_t,
\label{eq:latent_nudging}
\end{equation}
where $\eta$ is a constant step size. After updating $z_t$, we continue the standard diffusion process to generate the latent at $t{+}1$ (or the next index under the chosen scheduler) using the updated latent $z_t$ as the reference. We choose $\eta$ carefully in such a way that it does not destabilize the diffusion process. This form of test-time steering via gradient injection is consistent with prior diffusion guidance and posterior-sampling formulations, which demonstrate that differentiable objectives can be incorporated during sampling without retraining the generative model \cite{dhariwal2021diffusionbeatgans,chung2022dps}.

\medskip

For the implementation described in this paper, we use Stable Diffusion v1.5, since it natively supports inpainting in latent space \cite{rombach2021ldm}. We start by specifying a region of interest (ROI) in which we want to enforce a target color. Inpainting requires an initial image, so we generate a simple canvas image (via \texttt{matplotlib}) with a desired background color. We then replace the pixels within the target mask with zeros, producing an image with a ``hole.'' This image is passed through the diffusion inpainting pipeline, and the model attempts to fill only the masked area. Operationally, the model receives the complete image at each step (including the preserved background), meaning that the complete image undergoes the noising and denoising process: the model denoises the target region toward the prompt while simultaneously denoising the surrounding pixels back toward the original background. In practice, this induces an undesirable drift in the original color outside the mask. We additionally observe that because the complete image is denoised, the background color affects the generation within the target area. To combat drift in the background, we re-impose the latent in the preserved region by overwriting the non-masked area with a latent corresponding to the desired background color at each step.

\begin{figure}[H]
\centering
\includegraphics[width=\linewidth]{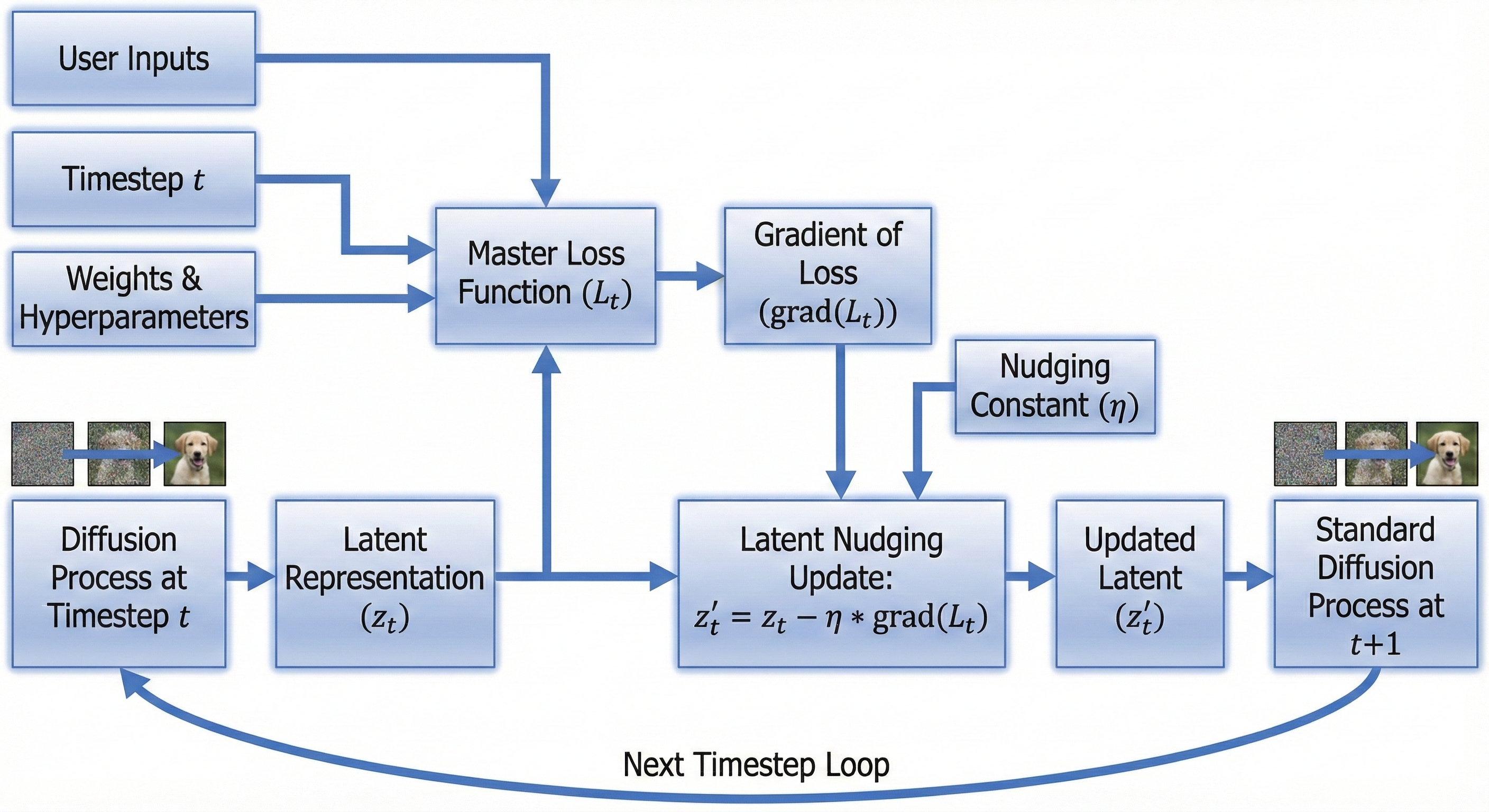}
\caption{\textbf{Loss function flow.} Diagram illustrating the inference-time optimization process and data flow.}
\label{fig:loss_flow}
\end{figure}

\begin{figure}[H]
\centering
\includegraphics[width=\linewidth]{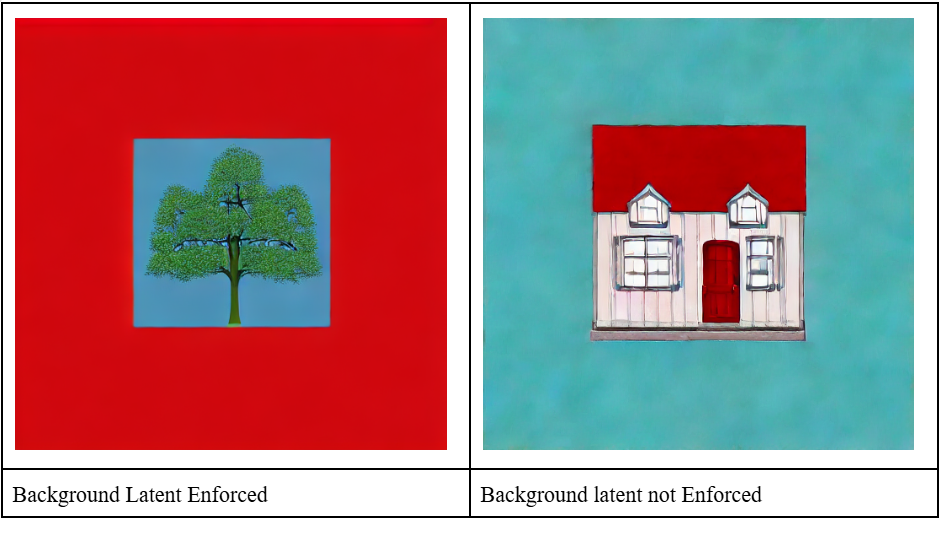}
\caption{\textbf{Background Consistency.} Notice the solid background colour when the latent is reinforced (left) versus the textured, inconsistent background when it is not (right).}
\label{fig:background_comparison}
\end{figure}

\medskip

Before introducing our final objective, we evaluated two simple mean-color losses as baselines. These losses were intentionally minimal: they provide an interpretable starting point for testing (i) inference-time latent nudging and (ii) the inpainting scheduler, but they do not attempt to control the full distribution of colors within the masked area.

\medskip

\medskip

Algorithm~1 computes the mean linear RGB color of all valid pixels in the masked area and penalizes its squared Euclidean deviation from the target color. Functionally, this loss enforces the \emph{average} color and is therefore effective at correcting gross hue biases. However, we did not adopt it as our primary objective because it is underconstrained: many visually distinct color fields share the same mean. In practice, this permits failure modes where the generation settles into a vague tint, or where a small subset of pixels becomes highly saturated (``spikes'') while the mean remains near the target. Additionally, as a purely mean-based constraint, it does not preserve fine-grained artifacts (e.g., enforcing the correct shade on leaves versus bark), since it provides no incentive for pixel-level consistency within the region.

\begin{algorithm}[H]
\caption{\texttt{LinearRGBEnforcerLoss}$(I,M,c^\star)$}
\begin{algorithmic}[1]
\State \textbf{Input:} linear RGB image $I\in\mathbb{R}^{3\times H\times W}$,
mask $M\in\{0,1\}^{H\times W}$,
target $c^\star\in\mathbb{R}^3$
\State Define valid pixels:
\[
\Omega=\{(h,w)\mid M_{h,w}=1,\ I_{:,h,w}\ \text{finite}\}
\]
\If{$|\Omega|=0$} \Return $0$ \EndIf
\State Compute mean color:
\[
\mu=\frac{1}{|\Omega|}\sum_{(h,w)\in\Omega} I_{:,h,w}
\]
\State \Return $\|\mu-c^\star\|_2^2$
\end{algorithmic}
\label{alg:linear_rgb}
\end{algorithm}

\medskip

Algorithm~2 mirrors the above approach, but operates in CIE Lab space and measures a channelwise Euclidean deviation between the mean Lab color and the target. This is preferable to RGB for perceptual reasons (Lab is approximately uniform), and empirically it reduces some hue-skew artifacts compared to RGB-only penalties. Nevertheless, we did not select this loss as the primary driver for the same structural reason: it remains a mean constraint and therefore cannot prevent heterogeneous solutions where a subset of pixels is far from the target while the mean is correct. In other words, it provides insufficient control over the \emph{tail} of the within-mask color error distribution, which is precisely where visually salient failures concentrate.

\begin{algorithm}[H]
\caption{\texttt{LabEuclideanLoss}$(I_{\text{lab}},M,c^\star_{\text{lab}})$}
\begin{algorithmic}[1]
\State \textbf{Input:} $I_{\text{lab}}\in\mathbb{R}^{3\times H\times W}$,
$M\in\{0,1\}^{1\times H\times W}$,
$c^\star_{\text{lab}}\in\mathbb{R}^3$
\State Cast $I_{\text{lab}},M,c^\star_{\text{lab}}$ to fp32
\State Reshape $c^\star_{\text{lab}}\rightarrow (1,3,1,1)$

\State Compute pixel count:
\[
n \gets \sum_{i,j} M_{:,i,j}
\]
\If{$n=0$} \Return $0$ \EndIf

\State Mask LAB values:
\[
I_{\Omega} \gets I_{\text{lab}}\odot M
\]

\State Compute mean LAB color:
\[
\mu \gets \frac{\sum_{i,j} I_{\Omega,:,i,j}}{n+\epsilon}
\]

\State Difference from target:
\[
d \gets \mu - c^\star_{\text{lab}}
\]

\State Channelwise Euclidean distance:
\[
\text{dist} \gets \sqrt{\sum_{c=1}^{3} d_c^2+\delta}
\]

\State \Return $\mathrm{mean}(\text{dist})$
\end{algorithmic}
\label{alg:lab_euclidean}
\end{algorithm}

\medskip
The mean-color baselines above are useful diagnostics, but they do not address the core failure mode in design-oriented generation: the model can satisfy a mean constraint while still producing locally incorrect color regions (e.g., a few leaves drifting to a wrong green, or small saturated patches that are perceptually salient). To prevent the model from converging to a vague tint and to preserve semantic artifacts, we adopt a pixelwise approach in the target region. The guiding principle is that if a specific shade is required for a particular object region, then the loss should (i) act on the mask at the pixel level and (ii) explicitly penalize both average deviation and worst-case deviation within that mask.

\medskip

Our final loss is a combination of multiple penalties derived from a per-pixel distance field in Lab space. The construction is modular: Algorithm~3 defines a per-pixel distance $u_t$ from the target color under an interpolation between a ``safe'' surrogate distance and a more direct perceptual squared distance. Algorithm~4 then produces a set of complementary penalties (mean constraint, pixelwise hinge, tail risk, soft maximum, and variance). Algorithm~5 combines these terms into a single scalar $L_t$ for use in inference-time latent nudging.

\medskip

We introduce a simple gating schedule $g(t)$ that increases with denoising progress. Intuitively, early timesteps are dominated by noise and the decoded image is not yet semantically stable; overly aggressive perceptual penalties at that stage can inject high-variance gradients. Our gate therefore implements a late start: guidance is weak or absent until a specified progress threshold, after which it ramps up smoothly. Operationally, this choice mirrors the empirical observation that enforcing strict perceptual constraints is most effective once the sample has entered a semantically meaningful regime.

\medskip

Finally, we emphasize that the composite loss is deliberately \emph{distribution-aware}. In addition to controlling the mean deviation, it penalizes (i) typical pixel deviations via $L_{\text{pix}}$, (ii) worst-case deviations via a tail-aggregation term (CVaR) and a smooth maximum (log-sum-exp), and (iii) within-mask dispersion via a variance constraint. This combination was selected to directly counteract the failure mode observed with mean-only objectives: small subsets of pixels that deviate substantially from the target while the mean remains acceptable. The CVaR component is included specifically to focus optimization pressure on the worst-performing fraction of pixels (tail risk) rather than treating all pixels uniformly \cite{rockafellar2000cvar}.

\medskip

\medskip

Algorithm~3 defines two distances: a channel-weighted surrogate $q_{\text{safe}}$ and a direct squared distance $q_{00}$. The surrogate weights luminance and chrominance differently (via $w_L$ and $w_{ab}$) to reduce instability in early guidance, while the direct distance increases perceptual strictness later in sampling. The gate $g$ interpolates between these distances, producing $q_t$, and the square-rooted field $u_t$ serves as a robust, differentiable per-pixel error magnitude. Algorithm~4 computes five penalties from $u_t$ and the Lab values: a mean constraint that enforces the average color up to a tolerance, a pixelwise hinge that discourages widespread per-pixel errors, a CVaR tail term that targets the largest errors (top $(1-\alpha)$ fraction), a smooth maximum based on log-sum-exp that approximates a max penalty without non-differentiability, and a variance term that discourages heterogeneous ``mixed'' solutions within the mask. Algorithm~5 ties these components together, applying the late-start gate as a function of denoising progress and returning the weighted sum $L_t$, which is then used as the inference-time guidance signal.

\begin{algorithm}[H]
\caption{\texttt{DistanceField}$(\ell,\mu,g)$}
\begin{algorithmic}[1]
\State \textbf{Input:} Lab image $\ell\in\mathbb{R}^{3\times H\times W}$,
target $\mu\in\mathbb{R}^3$,
gate $g\in[0,1]$
\State Define channel-weighted surrogate:
\[
q_{\text{safe}}(\ell,\mu)=w_L(\ell_L-\mu_L)^2+w_{ab}\|\ell_{ab}-\mu_{ab}\|_2^2
\]
\State Define perceptual squared distance:
\[
q_{00}(\ell,\mu)=\|\ell-\mu\|_2^2
\]
\State Gate interpolation:
\[
q_t=(1-g)\,q_{\text{safe}}+g\,q_{00}
\]
\State \Return distance field
\[
u_t=\sqrt{q_t+\epsilon}
\]
\end{algorithmic}
\label{alg:distance_field}
\end{algorithm}

\begin{algorithm}[H]
\caption{\texttt{ROILossTerms}$(u,\ell,\mu,M)$}
\begin{algorithmic}[1]
\State \textbf{Input:} distance field $u$, Lab image $\ell$, target $\mu$, mask $M$
\State ROI set $\Omega=\{i\mid M_i=1\}$, \quad $\phi_p(r)=\max(0,r)^p$

\State Mean Lab color:
\[
\hat{\mu}=\frac{1}{|\Omega|}\sum_{i\in\Omega}\ell_i
\]

\State Mean constraint:
\[
L_{\text{mean}}=\phi_p\big(\|\hat{\mu}-\mu\|_2-\tau_{\text{mean}}\big)
\]

\State Pixel-wise hinge penalty:
\[
L_{\text{pix}}=\frac{1}{|\Omega|}\sum_{i\in\Omega}\phi_p(u_i-\tau_{\text{pix}})
\]

\State Tail risk (CVaR):
\[
\mathrm{CVaR}_\alpha(u)=\frac{1}{k}\sum_{u_i\in\text{Top}(1-\alpha)}u_i
\]
\[
L_{\text{tail}}=\phi_p\big(\mathrm{CVaR}_\alpha(u)-\tau_{\text{tail}}\big)
\]

\State Soft maximum (LSE):
\[
\mathrm{LSE}_\beta(u)=\frac{1}{\beta}\log\Big(\tfrac{1}{|\Omega|}\sum_{i\in\Omega}e^{\beta u_i}\Big)
\]
\[
L_{\max}=\phi_p\big(\mathrm{LSE}_\beta(u)-\tau_{\max}\big)
\]

\State Variance constraint:
\[
\mathrm{Var}(u)=\frac{1}{|\Omega|}\sum_{i\in\Omega}(u_i-\bar{u})^2
\]
\[
L_{\text{var}}=\phi_p\big(\mathrm{Var}(u)-\tau_{\text{var}}\big)
\]

\State \Return $\{L_{\text{mean}},L_{\text{pix}},L_{\text{tail}},L_{\max},L_{\text{var}}\}$
\end{algorithmic}
\label{alg:roi_loss_terms}
\end{algorithm}

\begin{algorithm}[H]
\caption{\texttt{TotalLoss}$(x_t,t)$}
\begin{algorithmic}[1]
\State Decode and convert to Lab: $\ell\gets \mathrm{Lab}(\mathrm{VAE}(x_t))$

\State Compute denoising progress:
\[
\mathrm{prog}(t)=\frac{T-t}{T-t_{\min}}
\]

\State Late-start gate:
\[
g(t)=
\begin{cases}
0,& \mathrm{prog}(t)\le s\\[4pt]
\dfrac{\mathrm{prog}(t)-s}{1-s},& \mathrm{prog}(t)>s
\end{cases}
\]

\State Distance field:
\[
u_t\gets \texttt{DistanceField}(\ell,\mu,g(t))
\]

\State ROI penalty terms:
\[
(L_{\text{mean}},L_{\text{pix}},L_{\text{tail}},L_{\max},L_{\text{var}})
\gets \texttt{ROILossTerms}(u_t,\ell,\mu,M)
\]

\State Total loss:
\[
L_t=\lambda_{\text{mean}}L_{\text{mean}}
+\lambda_{\text{pix}}L_{\text{pix}}
+\lambda_{\text{tail}}L_{\text{tail}}
+\lambda_{\max}L_{\max}
+\lambda_{\text{var}}L_{\text{var}}
\]

\State \Return $L_t$
\end{algorithmic}
\label{alg:total_loss}
\end{algorithm}

We use a late start because the initial latents are highly noisy and can produce very large errors and unnecessary nudging of the latent. We find that while this loss does enforce the desired color as a mean, it fails to achieve the output we desire. This outcome is apparent visually, where small regions of the image become saturated deep blue. A further observation is that mentioning the target color in the prompt while using only this loss function can destabilize sampling, producing a failure case with an explosion in the loss values.

\medskip

Seeing the values of the losses and the nature of the output, we hypothesize that this might be because the loss becomes too strong toward the end of sampling. To validate this, we decayed the losses by multiplying the loss term by $1/t$, where $t$ is the timestep index.

\medskip

We observe a significant improvement in the quality of color enforcement in the output. Consistent with the decay, the logged loss trajectory exhibits a smoother decline. While this logging raises questions about whether optimization is concentrated earlier in sampling, we emphasize that the prompt does not mention the desired color at all. The constraint is induced entirely by the inference-time loss.

\medskip
\section*{Limitations}
Our method incurs additional inference-time compute due to per-step decoding, differentiable color-space conversion, and backpropagation through the loss to obtain guidance gradients; this overhead scales with the number of timesteps and the ROI size and can be prohibitive for interactive use at high resolutions. Performance is sensitive to mask quality: inaccurate ROI boundaries, soft edges, or segmentation leakage can cause color enforcement to spill into unintended regions or to under-enforce at the true boundary, especially when the object has fine structures. The approach relies on the fidelity and differentiability of the RGB-to-Lab conversion used during guidance; numerical approximations, gamut clipping, or implementation differences can bias gradients and create systematic color errors. Although the composite objective improves pixelwise adherence, stronger enforcement can trade off against texture realism and local detail, producing flattened chroma, banding, or unnatural transitions when the target color is inconsistent with the prompt-driven semantics. The method provides no formal guarantee of global optimum or constraint satisfaction: guidance strength, late-start thresholds, and schedule choices remain hyperparameter-dependent and can fail under difficult prompts, highly textured objects, extreme target colors, or strong prompt-color conflicts.

\medskip

\section*{Future Work}
A natural extension is to improve ROI boundary handling through boundary-aware regularization, alpha-matte masks, or explicit edge-consistency terms that reduce color bleeding while preserving fine structure. We also plan to replace fixed schedules with adaptive schedules that respond to denoising progress and loss curvature (e.g., increasing guidance only when the ROI error plateaus, and reducing it when gradients destabilize sampling). Another direction is to learn loss weights or thresholds from small calibration sets, enabling automatic tuning of $\lambda$ values and tolerances to balance adherence and realism for different prompt families and target colors. More broadly, incorporating richer color appearance models (beyond Lab), robust handling of gamut constraints, and illumination-aware objectives could reduce failure cases where perceptual color shifts arise from shading and highlights. For packaging and industrial design use-cases, we intend to integrate print-aware constraints such as ink-limit logic, $\Delta$E-based tolerances against reference swatches, and process priors (e.g., substrate and separation constraints) so that the guided outputs remain not only visually consistent on screen but also feasible under downstream reproduction pipelines.

\medskip

\bibliographystyle{plain}
\bibliography{library}

@misc{ho2020ddpm,
  title         = {Denoising Diffusion Probabilistic Models},
  author        = {Ho, Jonathan and Jain, Ajay and Abbeel, Pieter},
  year          = {2020},
  eprint        = {2006.11239},
  archivePrefix = {arXiv},
  primaryClass  = {cs.LG}
}

@misc{nichol2021improvedddpm,
  title         = {Improved Denoising Diffusion Probabilistic Models},
  author        = {Nichol, Alexander Quinn and Dhariwal, Prafulla},
  year          = {2021},
  eprint        = {2102.09672},
  archivePrefix = {arXiv},
  primaryClass  = {cs.LG}
}

@misc{song2020ddim,
  title         = {Denoising Diffusion Implicit Models},
  author        = {Song, Jiaming and Meng, Chenlin and Ermon, Stefano},
  year          = {2020},
  eprint        = {2010.02502},
  archivePrefix = {arXiv},
  primaryClass  = {cs.LG}
}

@misc{song2020sde,
  title         = {Score-Based Generative Modeling through Stochastic Differential Equations},
  author        = {Song, Yang and Sohl-Dickstein, Jascha and Kingma, Diederik P. and Kumar, Abhishek and Ermon, Stefano},
  year          = {2020},
  eprint        = {2011.13456},
  archivePrefix = {arXiv},
  primaryClass  = {cs.LG}
}

@misc{dhariwal2021diffusionbeatgans,
  title         = {Diffusion Models Beat {GAN}s on Image Synthesis},
  author        = {Dhariwal, Prafulla and Nichol, Alexander Quinn},
  year          = {2021},
  eprint        = {2105.05233},
  archivePrefix = {arXiv},
  primaryClass  = {cs.LG}
}

@misc{ho2022cfg,
  title         = {Classifier-Free Diffusion Guidance},
  author        = {Ho, Jonathan and Salimans, Tim},
  year          = {2022},
  eprint        = {2207.12598},
  archivePrefix = {arXiv},
  primaryClass  = {cs.LG}
}

@misc{rombach2021ldm,
  title         = {High-Resolution Image Synthesis with Latent Diffusion Models},
  author        = {Rombach, Robin and Blattmann, Andreas and Lorenz, Dominik and Esser, Patrick and Ommer, Bj{\"o}rn},
  year          = {2021},
  eprint        = {2112.10752},
  archivePrefix = {arXiv},
  primaryClass  = {cs.CV}
}

@misc{radford2021clip,
  title         = {Learning Transferable Visual Models From Natural Language Supervision},
  author        = {Radford, Alec and Kim, Jong Wook and Hallacy, Chris and Ramesh, Aditya and Goh, Gabriel and Agarwal, Sandhini and Sastry, Girish and Askell, Amanda and Mishkin, Pamela and Clark, Jack and Krueger, Gretchen and Sutskever, Ilya},
  year          = {2021},
  eprint        = {2103.00020},
  archivePrefix = {arXiv},
  primaryClass  = {cs.CV}
}

@inproceedings{devlin2019bert,
  title     = {{BERT}: Pre-training of Deep Bidirectional Transformers for Language Understanding},
  author    = {Devlin, Jacob and Chang, Ming-Wei and Lee, Kenton and Toutanova, Kristina},
  booktitle = {Proceedings of NAACL-HLT},
  year      = {2019},
  eprint    = {1810.04805},
  archivePrefix = {arXiv},
  primaryClass  = {cs.CL}
}

@misc{nichol2021glide,
  title         = {{GLIDE}: Towards Photorealistic Image Generation and Editing with Text-Guided Diffusion Models},
  author        = {Nichol, Alexander Quinn and Dhariwal, Prafulla and Ramesh, Aditya and Shyam, Pranav and Mishkin, Pamela and McGrew, Bob and Sutskever, Ilya and Chen, Mark},
  year          = {2021},
  eprint        = {2112.10741},
  archivePrefix = {arXiv},
  primaryClass  = {cs.CV}
}

@misc{butt2025gencolorbench,
  title         = {GenColorBench: Improving Color Fidelity in Text-to-Image Generation},
  author        = {Butt, Muhammad Atif and Gomez-Villa, Alexandra and Wu, Tao and Vazquez-Corral, Javier and Van De Weijer, Joost and Wang, Kai},
  year          = {2025},
  eprint        = {2510.20586},
  archivePrefix = {arXiv},
  primaryClass  = {cs.CV}
}

@misc{zhang2023controlnet,
  title         = {Adding Conditional Control to Text-to-Image Diffusion Models},
  author        = {Zhang, Lvmin and Rao, Anyi and Agrawala, Maneesh},
  year          = {2023},
  eprint        = {2302.05543},
  archivePrefix = {arXiv},
  primaryClass  = {cs.CV}
}

@misc{cao2024controllablesurvey,
  title         = {Controllable Generation with Text-to-Image Diffusion Models: A Survey},
  author        = {Cao, Pu and Zhou, Feng and Song, Qing and Yang, Lu},
  year          = {2024},
  eprint        = {2403.04279},
  archivePrefix = {arXiv},
  primaryClass  = {cs.CV}
}

@misc{vavilala2023palette_local_control,
  title         = {Applying a Color Palette with Local Control using Diffusion Models},
  author        = {Vavilala, Vaibhav and Forsyth, David},
  year          = {2023},
  eprint        = {2307.02698},
  archivePrefix = {arXiv},
  primaryClass  = {cs.CV}
}

@misc{aharoni2025palette_aligned,
  title         = {Palette-Aligned Image Diffusion},
  author        = {Aharoni, Ido and Levi, Omer and Hassid, Shai and Michaeli, Tomer},
  year          = {2025},
  eprint        = {2509.02000},
  archivePrefix = {arXiv},
  primaryClass  = {cs.CV}
}

@misc{agarwal2024trainingfree_colorstyle,
  title         = {Training-free Color-Style Disentanglement for Constrained Text-to-Image Synthesis},
  author        = {Agarwal, Aishwarya and Karanam, Srikrishna and Vasan Srinivasan, Balaji},
  year          = {2024},
  eprint        = {2409.02429},
  archivePrefix = {arXiv},
  primaryClass  = {cs.CV}
}

@misc{chung2022dps,
  title         = {Diffusion Posterior Sampling for General Noisy Inverse Problems},
  author        = {Chung, Hyungjin and Kim, Jeongsol and McCann, Michael T. and Klasky, Marc L. and Ye, Jong Chul},
  year          = {2022},
  eprint        = {2209.14687},
  archivePrefix = {arXiv},
  primaryClass  = {cs.CV}
}

@misc{kawar2022ddrm,
  title         = {Denoising Diffusion Restoration Models},
  author        = {Kawar, Bahjat and Elad, Michael and Ermon, Stefano and Song, Jiaming},
  year          = {2022},
  eprint        = {2201.11793},
  archivePrefix = {arXiv},
  primaryClass  = {cs.CV}
}

@misc{wang2022ddnm,
  title         = {Denoising Diffusion Null-Space Model for Inverse Problems},
  author        = {Wang, Yinhuai and Yu, Jiwen and Zhang, Jian},
  year          = {2022},
  eprint        = {2212.00490},
  archivePrefix = {arXiv},
  primaryClass  = {cs.CV}
}

@inproceedings{lobashev2025sw_guidance,
  title     = {Color Conditional Generation with Sliced-Wasserstein Guidance},
  author    = {Lobashev, Alexander and Larchenko, Maria and Guskov, Dmitry},
  booktitle = {Advances in Neural Information Processing Systems (NeurIPS)},
  year      = {2025},
  note      = {Poster}
}

@inproceedings{riba2020kornia,
  author    = {Riba, Edgar and Mishkin, Dmytro and Ponsa, Daniel and Rublee, Ethan and Bradski, Gary},
  title     = {Kornia: an Open Source Differentiable Computer Vision Library for PyTorch},
  booktitle = {Proceedings of the IEEE/CVF Winter Conference on Applications of Computer Vision (WACV)},
  year      = {2020},
  pages     = {3674--3683}
}

@misc{cie11664_4_2019,
  author       = {{International Organization for Standardization} and {International Commission on Illumination (CIE)}},
  title        = {Colorimetry --- Part 4: CIE 1976 L*a*b* colour space (ISO/CIE 11664-4:2019)},
  year         = {2019},
  note         = {Joint ISO/CIE Standard}
}

@misc{icc_srgb,
  author       = {{International Color Consortium (ICC)}},
  title        = {sRGB Color Space Profile (sRGB2014 and related profile/transfer specification materials)},
  year         = {2014},
  note         = {Accessed via ICC profile registry}
}

@article{rockafellar2000cvar,
  title   = {Optimization of Conditional Value-at-Risk},
  author  = {Rockafellar, R. Tyrrell and Uryasev, Stanislav},
  journal = {Journal of Risk},
  year    = {2000},
  volume  = {2},
  number  = {3},
  pages   = {21--41}
}

\newpage
\section*{Appendices}

\subsection*{Appendix 1: Second-Order Taylor Approximation of $\Delta E_{00}$}
\begin{algorithm}[H]
\caption{Second-Order Taylor Approximation of $\Delta E_{00}$}
\label{alg:taylor_approximation}
\begin{algorithmic}[1]
\State \textbf{Input:} Lab color $\ell_1 = (L_1,a_1,b_1)$, $\ell_2 = (L_2,a_2,b_2)$
\State Define differences:
\[
\Delta \ell = \ell_1 - \ell_2, \quad \Delta L = L_1 - L_2, \quad \Delta a = a_1 - a_2, \quad \Delta b = b_1 - b_2.
\]
\State Local second-order expansion:
\[
\Delta E_{00}^{\,2}
= (\Delta L)^2 + (\Delta a)^2 + (\Delta b)^2
+ O\!\left(\|\Delta \ell\|^{3}\right).
\]
\State \textbf{Result:} Second-order Taylor approximation
\[
\boxed{\Delta E_{00} \approx \sqrt{(\Delta L)^2 + (\Delta a)^2 + (\Delta b)^2}}
\]
\State Safe quadratic form:
\[
\boxed{Q = (\Delta L)^2 + (\Delta a)^2 + (\Delta b)^2.}
\]
\end{algorithmic}
\end{algorithm}
\vspace{1em}

\subsection*{Appendix 2: Inpainting Diffusion Algorithm}
\begin{algorithm}[H]
\caption{Inpainting Diffusion with Inference-Time Color Guidance (Latent Nudging) --- Part I}
\label{alg:color_guided_inpaint_appendix_part1}
\begin{algorithmic}[1]
\Require Pipeline $\mathcal{P}=(\mathrm{VAE},\mathrm{UNet},\mathrm{Scheduler})$; canvas $X\in[0,1]^{3\times H\times W}$; ROI mask $M\in\{0,1\}^{1\times H\times W}$; embeddings $(e_{\mathrm{cond}},e_{\mathrm{uncond}})$; steps $N$; CFG scale $s_{\mathrm{cfg}}$; nudge step $\eta$; master weight $\lambda$; window fractions $(f_{\mathrm{start}},f_{\mathrm{stop}})$; toggles $\mathsf{LinRGB},\mathsf{CVaR},\mathsf{Anchor}$; target color $c^\star_{\mathrm{sRGB}}\in[0,1]^3$; hyperparameters $\Theta$.
\Ensure Final latent $z$ and logs (losses, grad norms, ROI means, snapshots).
\medskip

\State \textbf{Scheduler:} $\mathrm{Scheduler.set\_timesteps}(N)$; $\{t_i\}_{i=0}^{N-1}\leftarrow \mathrm{Scheduler.timesteps}$.
\State $i_{\mathrm{start}}\!\leftarrow\!\lfloor f_{\mathrm{start}}N\rfloor$, \; $i_{\mathrm{stop}}\!\leftarrow\!\lfloor f_{\mathrm{stop}}N\rfloor$.
\State \textbf{Encode canvas:} $z_0\leftarrow \alpha_{\mathrm{vae}}\cdot \mathrm{VAE.encode}(X)$, \; $\alpha_{\mathrm{vae}}=0.18215$.
\State \textbf{Latent mask:} $\tilde M\leftarrow \mathrm{Resize}(M,\mathrm{shape}(z_0)_{H,W})$.
\State \textbf{Masked-image latents:} $X_{\mathrm{mask}}\leftarrow X\odot(1-M)$, \; $z_{\mathrm{mask}}\leftarrow \alpha_{\mathrm{vae}}\cdot \mathrm{VAE.encode}(X_{\mathrm{mask}})$.
\State \textbf{Init noise/latents:} $\epsilon\sim\mathcal{N}(0,I)$, \; $z\leftarrow \sigma_0\epsilon$, \; $\sigma_0=\mathrm{Scheduler.init\_noise\_sigma}$.
\State \textbf{Targets (once):} $c^\star_{\mathrm{lin}}\leftarrow \mathrm{sRGB}\!\to\!\mathrm{LinRGB}(c^\star_{\mathrm{sRGB}})$, \;
$c^\star_{\mathrm{lab}}\leftarrow \mathrm{sRGB}\!\to\!\mathrm{Lab}(c^\star_{\mathrm{sRGB}})$.
\State $T_0\leftarrow t_0$.
\medskip

\For{$i=0$ \textbf{to} $N-1$}
  \State $t\leftarrow t_i$.
  \medskip

  \State \textbf{(1) \textbf{Inpainting denoise (CFG):}}
  \[
  \hat z \leftarrow \mathrm{Scheduler.scale\_model\_input}([z;z],t),\quad
  u \leftarrow [\hat z;\,[\tilde M;\tilde M];\,[z_{\mathrm{mask}};z_{\mathrm{mask}}]].
  \]
  \[
  \hat\epsilon \leftarrow \mathrm{UNet}(u,t,[e_{\mathrm{uncond}};e_{\mathrm{cond}}]),\quad
  \hat\epsilon=\hat\epsilon_u\oplus \hat\epsilon_c,\quad
  \hat\epsilon_{\mathrm{cfg}} \leftarrow \hat\epsilon_u+s_{\mathrm{cfg}}(\hat\epsilon_c-\hat\epsilon_u).
  \]
  \[
  z \leftarrow \mathrm{Scheduler.step}(\hat\epsilon_{\mathrm{cfg}},t,z).
  \]
  \medskip

  \If{$\mathsf{Anchor}$}
    \State \textbf{(Optional) Background anchor:}
    \[
    z_{\mathrm{bg}}(t)\leftarrow \mathrm{Scheduler.add\_noise}(z_0,\epsilon,t),\qquad
    z\leftarrow (1-\tilde M)\odot z_{\mathrm{bg}}(t)+\tilde M\odot z.
    \]
  \EndIf
  \medskip

  \State \textbf{(2) Guidance gate:}
  \[
  \mathsf{apply}\leftarrow (i\ge i_{\mathrm{start}})\wedge(i<i_{\mathrm{stop}}),\qquad
  \mathsf{on}\leftarrow \mathsf{LinRGB}\vee \mathsf{CVaR}.
  \]
  \If{\textbf{not} $\mathsf{apply}$ \textbf{or} \textbf{not} $\mathsf{on}$}
    \State Record zeros; \textbf{continue}.
  \EndIf
\algstore{myalg}
\end{algorithmic}
\end{algorithm}

\begin{algorithm}[H]
\caption{Inpainting Diffusion with Inference-Time Color Guidance (Latent Nudging) --- Part II}
\label{alg:color_guided_inpaint_appendix_part2}
\begin{algorithmic}[1]
\algrestore{myalg}
  \State \textbf{(3) Decode + ROI means (monitoring):}
  \[
  x\leftarrow \mathrm{VAE.decode}(z/\alpha_{\mathrm{vae}}),\quad
  \ell\leftarrow \mathrm{VAE}\!\to\!\mathrm{Lab}(x),\quad
  r\leftarrow \mathrm{VAE}\!\to\!\mathrm{sRGB}(x),\quad
  n\leftarrow \sum_{h,w} M_{1,h,w}.
  \]
  \[
  \bar\ell\leftarrow \frac{1}{n+\varepsilon}\sum_{h,w}\ell_{:,h,w}M_{1,h,w},\qquad
  \bar r\leftarrow \frac{1}{n+\varepsilon}\sum_{h,w}r_{:,h,w}M_{1,h,w}.
  \]
  \State Append $(\bar\ell,\bar r)$ and optional snapshots.
  \medskip

  \State \textbf{(4) Losses + schedules:} $L_{\mathrm{lin}}\leftarrow 0$, $L_{\mathrm{cvar}}\leftarrow 0$.
  \If{$\mathsf{LinRGB}$}
    \[
    x_{\mathrm{lin}}\leftarrow \mathrm{VAE}\!\to\!\mathrm{LinRGB}(x),\quad
    w_{\mathrm{lin}}(i)\leftarrow \frac{1}{i+1},
    \quad
    L_{\mathrm{lin}}\leftarrow w_{\mathrm{lin}}(i)\,\lambda_{\mathrm{lin}}\,
    \mathcal{L}_{\mathrm{LinRGB}}(x_{\mathrm{lin}},M,c^\star_{\mathrm{lin}}).
    \]
  \EndIf
  \medskip
  \If{$\mathsf{CVaR}$ \textbf{and} $k\neq 0$}
    \[
    w_{\mathrm{cvar}}(t)\leftarrow \frac{\max(0,T_0-t)}{k},\qquad
    L_{\mathrm{cvar}}\leftarrow w_{\mathrm{cvar}}(t)\,
    \mathcal{L}_{\mathrm{CVaR}}(x,t,\{t_i\},f_{\mathrm{start}},c^\star_{\mathrm{lab}},M;\Theta).
    \]
  \EndIf
  \medskip
  \[
  L \leftarrow \lambda\,(L_{\mathrm{lin}}+L_{\mathrm{cvar}}).
  \]
  \State Record $L_{\mathrm{lin}},L_{\mathrm{cvar}}$ and (if available) CVaR sub-terms.
  \medskip

  \State \textbf{(5) Latent nudging (ROI-only):}
  \If{\textbf{not} $\mathrm{finite}(L)$ \textbf{or} $L\le 0$}
    \State Record zeros; \textbf{continue}.
  \EndIf
  \State Compute gradient $g\leftarrow \nabla_z L$; $g\leftarrow \mathrm{clip}(g,-1,1)$; $a\leftarrow \|g\|_2$.
  \If{$a\le 0$} \State Record and \textbf{continue}. \EndIf
  \[
  \tilde g \leftarrow \frac{g}{a+\varepsilon},\qquad
  z \leftarrow z - \eta\,\tilde g \odot \tilde M .
  \]
  \State Replace NaN/Inf if present; record $L$ and $a$.
\EndFor

\State \Return final $z$ with logs and runtime.
\end{algorithmic}
\end{algorithm}

\subsection*{Appendix 3: Experimental Results (CVaR + L-RGB)}
\label{app:results}

We present the experimental validation of our proposed CVaR + Linear-RGB loss configuration. The results demonstrate the effectiveness of the composite objective in steering the latent generation toward the target color within the Region of Interest (ROI).

\subsubsection*{Run Summary}
\begin{itemize}
    \item \textbf{Device:} Tesla T4 (Runtime: 39.9s)
    \item \textbf{Steps:} 80 (Enforcer active: 20\%--100\%)
    \item \textbf{Parameters:} $\eta=0.009$, $\lambda_{\text{guidance}}=100$, $\lambda_{\text{master}}=0.07$, $s_{\text{cfg}}=8$, $k=2$
    \item \textbf{Toggles:} LinearRGB=True, Angad=True, BgAnchor=True
\end{itemize}

\subsubsection*{Target vs. Final Color}
\begin{table}[H]
    \centering
    \begin{tabular}{l l l l}
        \hline
        \textbf{Metric} & \textbf{L} & \textbf{a} & \textbf{b} \\
        \hline
        Target (Lab) & 69.97 & 47.40 & 11.54 \\
        Target (sRGB) & \multicolumn{3}{l}{[1.00, 0.53, 0.60]} \\
        Final (Lab) & 62.71 & 25.60 & -6.48 \\
        \hline
        \textbf{Error ($\Delta E_{00}$)} & \multicolumn{3}{l}{\textbf{29.19} (Mean over ROI)} \\
        \hline
    \end{tabular}
    \caption{Target vs. Achieved Color Statistics}
\end{table}

\subsubsection*{Pixelwise $\Delta E_{00}$ Statistics}
Analysis of pixelwise error distribution across the ROI:
\begin{itemize}
    \item \textbf{Min:} 0.36
    \item \textbf{Max:} 72.05
    \item \textbf{Mean:} 52.95
    \item \textbf{Std Dev:} 19.36
    \item \textbf{Median:} 62.68
\end{itemize}

\subsubsection*{Visual Analysis}
\begin{figure}[H]
    \centering
    \includegraphics[width=\linewidth]{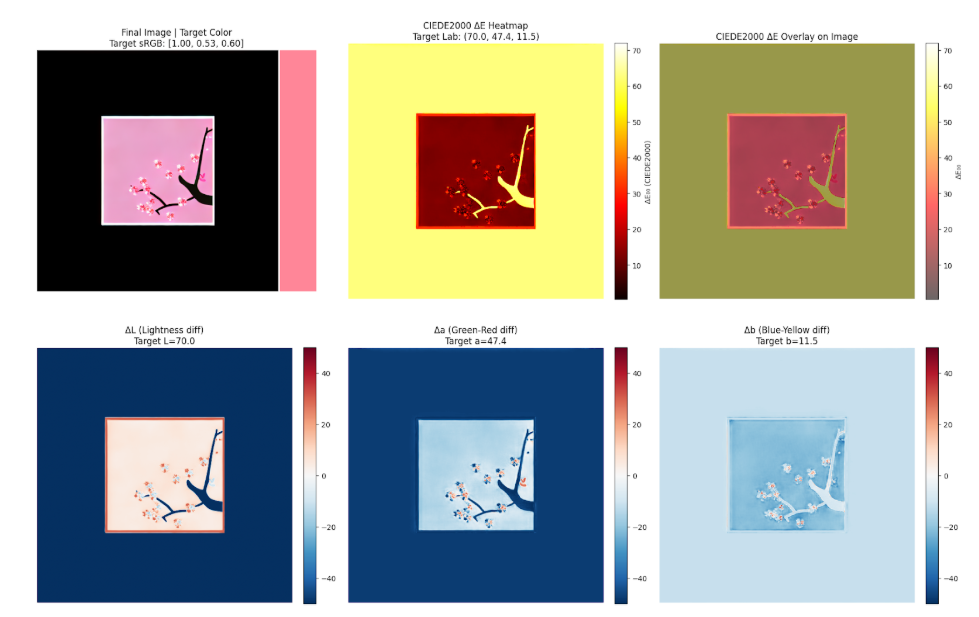}
    \caption{\textbf{Qualitative Results and Error Heatmaps.} Top row: Final image vs. target swatch, $\Delta E_{00}$ heatmap, and overlay. Bottom row: Component difference maps for $\Delta L$, $\Delta a$, and $\Delta b$.}
    \label{fig:cvar_heatmaps}
\end{figure}

\begin{figure}[H]
    \centering
    \includegraphics[width=\linewidth]{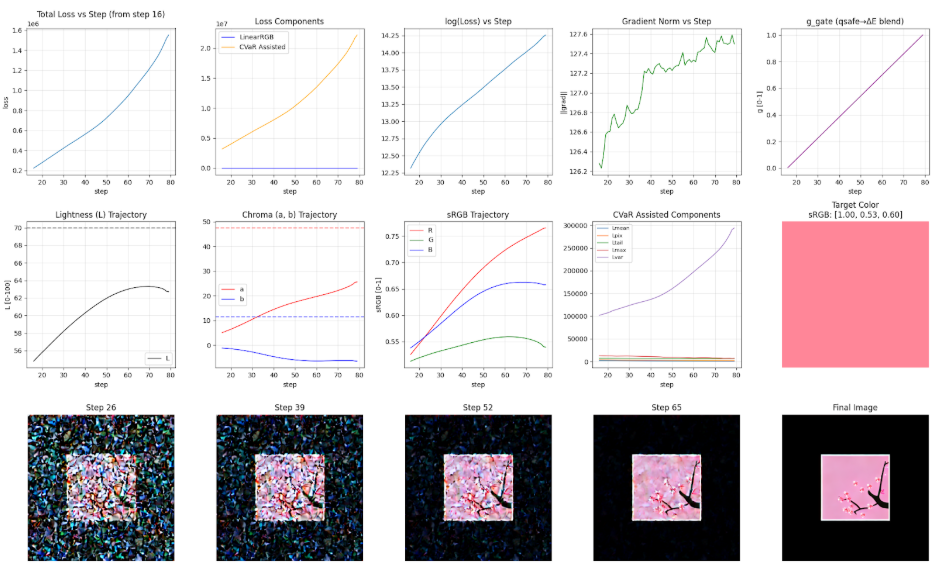}
    \caption{\textbf{Loss and Optimization Dynamics.} Evolution of total loss, component losses (LinearRGB vs. CVaR), gradient norms, and sRGB trajectory over 80 sampling steps.}
    \label{fig:cvar_graphs}
\end{figure}

\begin{figure}[H]
    \centering
    \includegraphics[width=\linewidth]{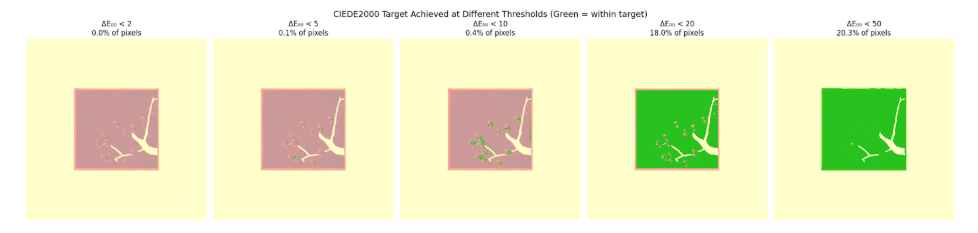}
    \caption{\textbf{Threshold Analysis.} Visualization of ROI pixels satisfying $\Delta E_{00} < T$ for thresholds $T \in \{2, 5, 10, 20, 50\}$. Green indicates satisfaction.}
    \label{fig:cvar_thresholds}
\end{figure}

Figure \ref{fig:cvar_heatmaps} illustrates the final output compared to the target, along with a $\Delta E_{00}$ heatmap showing the spatial distribution of error. Figure \ref{fig:cvar_graphs} details the loss trajectory, gradient norms, and component sub-losses throughout the sampling process. Figure \ref{fig:cvar_thresholds} analyzes the percentage of ROI pixels achieving various $\Delta E_{00}$ thresholds.

\subsection*{Appendix 4: Experimental Results (Linear-RGB MSE)}
\label{app:results_lrgb}

We additionally present results for a configuration using only the Linear-RGB MSE loss (without the distribution-aware CVaR term). This baseline demonstrates the limitations of mean-based objectives in capturing perceptual color accuracy.

\subsubsection*{Run Summary}
\begin{itemize}
    \item \textbf{Device:} Tesla T4 (Runtime: 37.8s)
    \item \textbf{Steps:} 80 (Enforcer active: 20\%--100\%)
    \item \textbf{Parameters:} $\eta=0.009$, $\lambda_{\text{guidance}}=100$, $\lambda_{\text{master}}=0.07$, $s_{\text{cfg}}=8$, $k=2$
    \item \textbf{Toggles:} LinearRGB=True, Angad=False, BgAnchor=True
\end{itemize}

\subsubsection*{Target vs. Final Color}
\begin{table}[H]
    \centering
    \begin{tabular}{l l l l}
        \hline
        \textbf{Metric} & \textbf{L} & \textbf{a} & \textbf{b} \\
        \hline
        Target (Lab) & 69.97 & 47.40 & 11.54 \\
        Target (sRGB) & \multicolumn{3}{l}{[1.00, 0.53, 0.60]} \\
        Final (Lab) & 55.67 & 17.32 & -7.26 \\
        \hline
        \textbf{Error ($\Delta E_{00}$)} & \multicolumn{3}{l}{\textbf{38.24} (Mean over ROI)} \\
        \hline
    \end{tabular}
    \caption{Target vs. Achieved Color Statistics (Linear-RGB only)}
\end{table}

\subsubsection*{Pixelwise $\Delta E_{00}$ Statistics}
Analysis of pixelwise error distribution across the ROI:
\begin{itemize}
    \item \textbf{Min:} 3.87
    \item \textbf{Max:} 68.50
    \item \textbf{Mean:} 55.12
    \item \textbf{Std Dev:} 15.41
    \item \textbf{Median:} 62.68
\end{itemize}

\subsubsection*{Visual Analysis}
\begin{figure}[H]
    \centering
    \includegraphics[width=\linewidth]{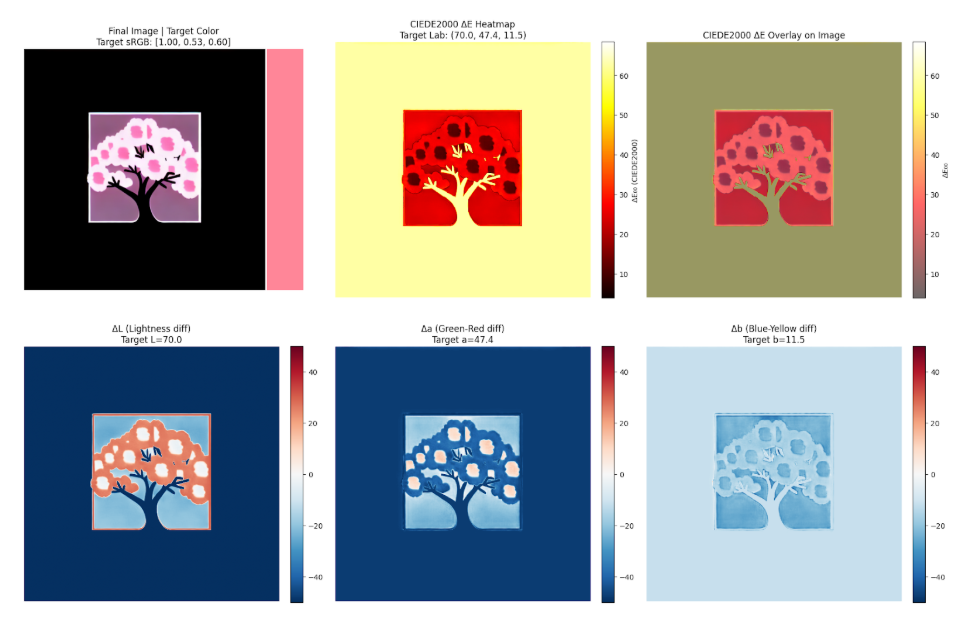}
    \caption{\textbf{Qualitative Results (Linear-RGB Best Effort).} Note the reduced saturation and shift in hue compared to the CVaR-guided result.}
    \label{fig:lrgb_heatmaps}
\end{figure}

\begin{figure}[H]
    \centering
    \includegraphics[width=\linewidth]{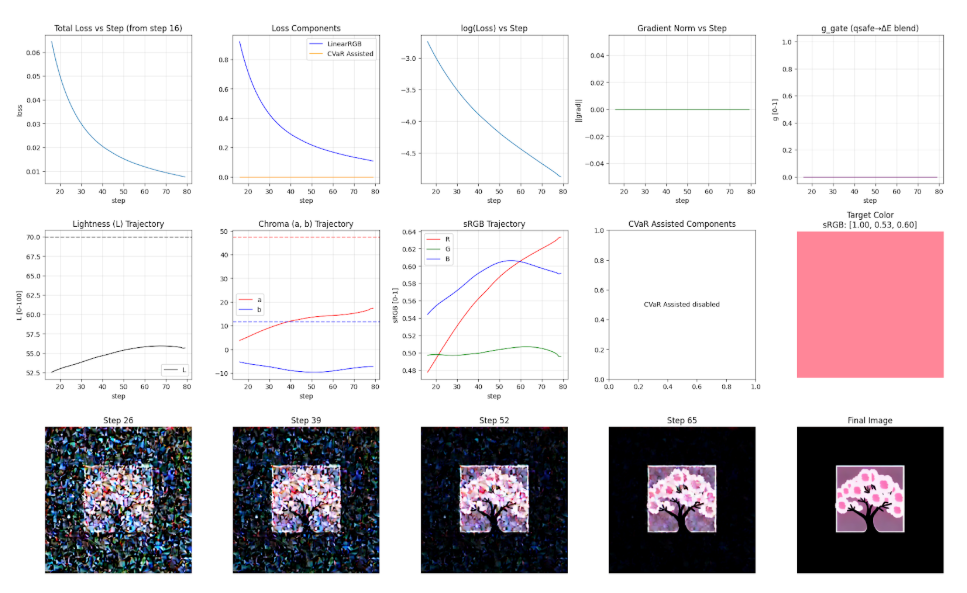}
    \caption{\textbf{Loss Dynamics (Linear-RGB Only).} The loss decreases smoothly but fails to enforce the strict perceptual constraint.}
    \label{fig:lrgb_graphs}
\end{figure}

\begin{figure}[H]
    \centering
    \includegraphics[width=\linewidth]{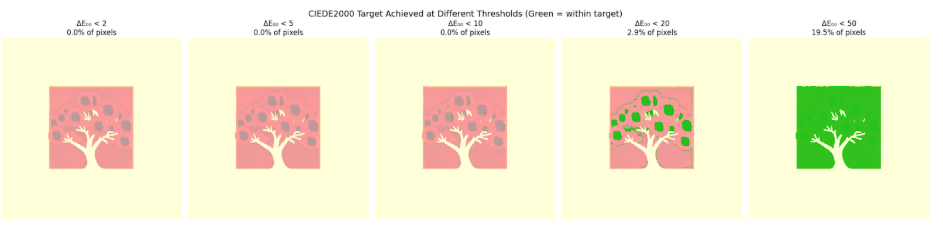}
    \caption{\textbf{Threshold Analysis (Linear-RGB).} Satisfaction rates are significantly lower than the CVaR-assisted run, with 0\% of pixels meeting the $\Delta E_{00} < 10$ threshold.}
    \label{fig:lrgb_thresholds}
\end{figure}

Figure \ref{fig:lrgb_heatmaps} shows the qualitative results, where visually significant color deviation is apparent compared to the target. Figure \ref{fig:lrgb_graphs} shows the loss evolution, and Figure \ref{fig:lrgb_thresholds} highlights the threshold satisfaction rates.

\subsection*{Appendix 5: Experimental Results (No-Guidance Baseline)}
\label{app:results_noloss}

Finally, we present results for a configuration with no inference-time guidance enabled. This serves as a negative control to highlight the necessity of active guidance for color enforcement.

\subsubsection*{Run Summary}
\begin{itemize}
    \item \textbf{Device:} Tesla T4 (Runtime: 11.6s)
    \item \textbf{Steps:} 80 (Enforcer active: 20\%--100\%)
    \item \textbf{Parameters:} $\eta=0.009$, $\lambda_{\text{guidance}}=100$, $\lambda_{\text{master}}=0.07$, $s_{\text{cfg}}=8$, $k=2$
    \item \textbf{Toggles:} LinearRGB=False, Angad=False, BgAnchor=True
\end{itemize}

\subsubsection*{Target vs. Final Color}
\begin{table}[H]
    \centering
    \begin{tabular}{l l l l}
        \hline
        \textbf{Metric} & \textbf{L} & \textbf{a} & \textbf{b} \\
        \hline
        Target (Lab) & 69.97 & 47.40 & 11.54 \\
        Target (sRGB) & \multicolumn{3}{l}{[1.00, 0.53, 0.60]} \\
        Final (Lab) & 52.87 & 21.41 & -11.49 \\
        \hline
        \textbf{Error ($\Delta E_{00}$)} & \multicolumn{3}{l}{\textbf{38.71} (Mean over ROI)} \\
        \hline
    \end{tabular}
    \caption{Target vs. Achieved Color Statistics (No Guidance)}
\end{table}

\subsubsection*{Pixelwise $\Delta E_{00}$ Statistics}
Analysis of pixelwise error distribution across the ROI:
\begin{itemize}
    \item \textbf{Min:} 5.65
    \item \textbf{Max:} 73.28
    \item \textbf{Mean:} 55.22
    \item \textbf{Std Dev:} 16.84
    \item \textbf{Median:} 62.68
\end{itemize}

\subsubsection*{Visual Analysis}
\begin{figure}[H]
    \centering
    \includegraphics[width=\linewidth]{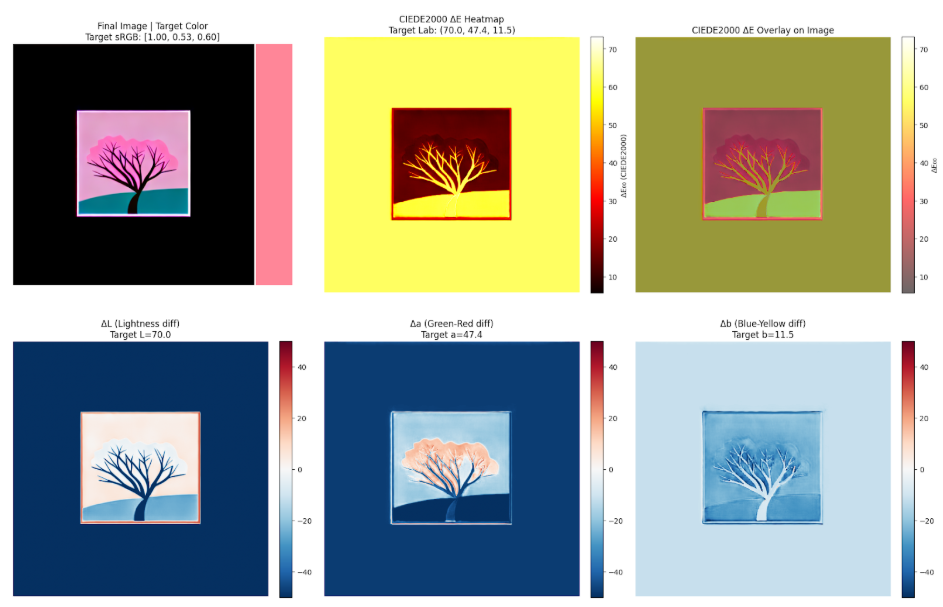}
    \caption{\textbf{Qualitative Results (No Guidance).} The output color is generic and fails to match the specific target swatch.}
    \label{fig:noloss_heatmaps}
\end{figure}

\begin{figure}[H]
    \centering
    \includegraphics[width=\linewidth]{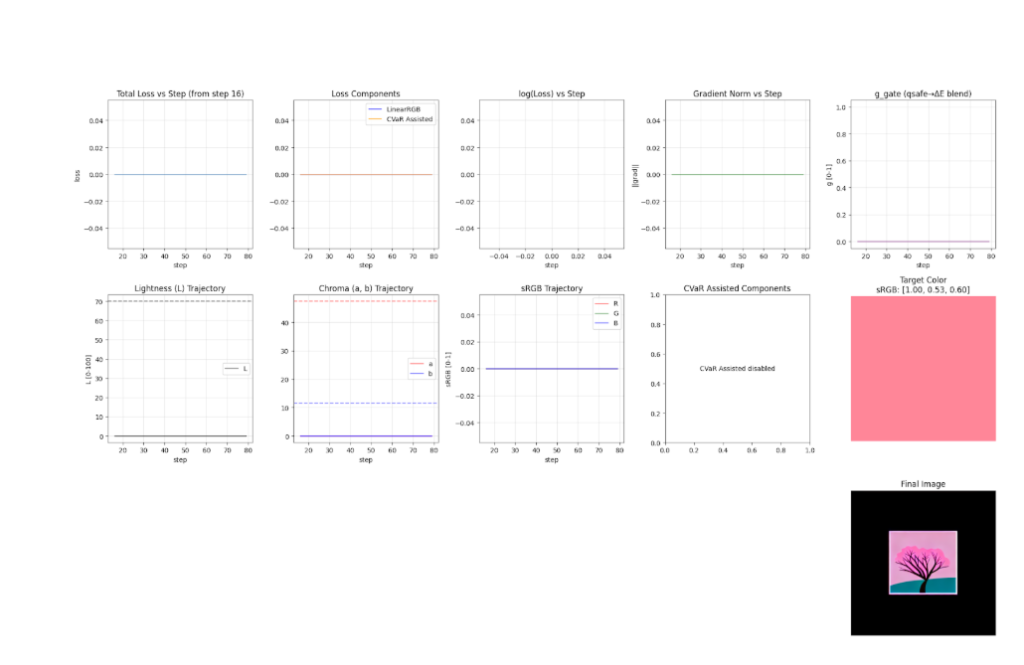}
    \caption{\textbf{Loss Dynamics (No Guidance).} Flat lines indicate that no guidance loss was computed or applied.}
    \label{fig:noloss_graphs}
\end{figure}

\begin{figure}[H]
    \centering
    \includegraphics[width=\linewidth]{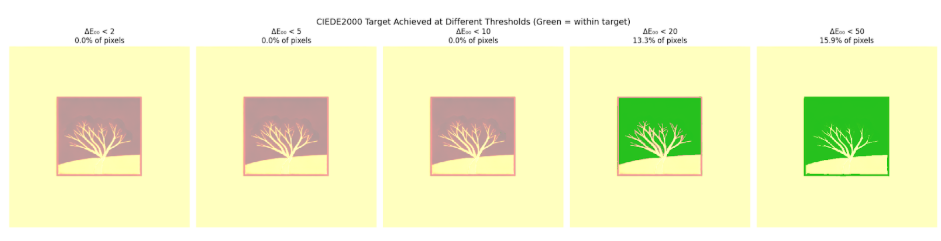}
    \caption{\textbf{Threshold Analysis (No Guidance).} The vast majority of pixels fall into the highest error accumulation bin ($\Delta E_{00} \ge 20$).}
    \label{fig:noloss_thresholds}
\end{figure}

Figure \ref{fig:noloss_heatmaps} illustrates the baseline performance. Without guidance, the model generates a generic pink that drifts significantly from the target. Figure \ref{fig:noloss_graphs} shows the lack of active loss components. Figure \ref{fig:noloss_thresholds} confirms that nearly 87\% of pixels have a $\Delta E_{00} \ge 20$, indicating a "very different" color.

\end{document}